\documentclass{article}
\usepackage[english]{babel}
\usepackage[utf8]{inputenc}
\usepackage{xcolor}

\usepackage{johd}

\title{Neural Language Models for Nineteenth-Century English}

\author{Kasra Hosseini$^{a}$$^{*}$, 
        Kaspar Beelen$^{a}$$^{,}$$^{b}$, 
        Giovanni Colavizza$^{c}$, 
        Mariona Coll Ardanuy$^{a}$$^{,}$$^{b}$ \\
        \small $^{a}$The Alan Turing Institute, London, UK \\
        \small $^{b}$Queen Mary University of London, London, UK \\
        \small $^{c}$Institute for Logic, Language and Computation, University of Amsterdam, NL\\
        \\
        \small $^{*}$Corresponding author: Kasra Hosseini; \tt{khosseini@turing.ac.uk} \\
}
\date{}

\begin{document}
\maketitle

% =================== 
\begin{abstract} 
\noindent
We present four types of neural language models trained on a large historical dataset of books in English, published between 1760-1900 and comprised of $\approx$5.1 billion tokens. The language model architectures include static (word2vec and fastText) and contextualized models (BERT and Flair). For each architecture, we trained a model instance using the whole dataset. Additionally, we trained separate instances on text published before 1850 for the two static models, and four instances considering different time slices for BERT. Our models have already been used in various downstream tasks where they consistently improved performance. In this paper, we describe how the models have been created and outline their reuse potential.
\end{abstract}

\noindent\keywords{language model; BERT; word2vec; fastText; nineteenth-century English; digital heritage.}\\

% =================== 
\section{Overview}

As language is subject to continuous change, the computational analysis of digital (textual) heritage should attune models and methods to the specific historical contexts in which these texts emerged.
This paper aims to facilitate the ``historicization'' of Natural Language Processing (NLP) methods by releasing various language models trained on a 19th-century book collection.
These models can support research in digital and computational humanities, history, computational linguistics and the cultural heritage or GLAM sector (galleries, libraries, archives, and museums).
To accommodate different research needs, we release a wide variety of models, from ``static'' embeddings (word2vec and fastText) to more recent language models that produce context-dependent word or string embeddings (BERT and Flair, respectively). Here, we consider a model ``static'' when it generates only one embedding for a given token at inference time, regardless of the textual context in which the token appears. On the other hand, ``contextual'' models generate a distinct embedding according to the textual context at inference time.

\paragraph{Repository location} The dataset is available on Zenodo at \url{http://doi.org/10.5281/zenodo.4782245}.

\paragraph{Context} This work was produced as part of Living with Machines (LwM),\footnote{\url{https://livingwithmachines.ac.uk} (last access: 2021-05-24).} an interdisciplinary project focused on the lived experience of Britain's industrialization during the long 19th century. The language models presented here have been used in several research projects, to assess the impact of optical character recognition (OCR) on NLP tasks~\citep{van2020assessing}, to detect atypical animacy~\citep{coll-ardanuy-etal-2020-living}, and for targeted sense disambiguation~\citep{beelen-etal-2021-tsd}.

% ===================
\section{Method}

% ===================
\subsection{Original corpus}

% ===================
\begin{figure}
	\begin{center}
		\includegraphics[scale=0.80, angle=0]{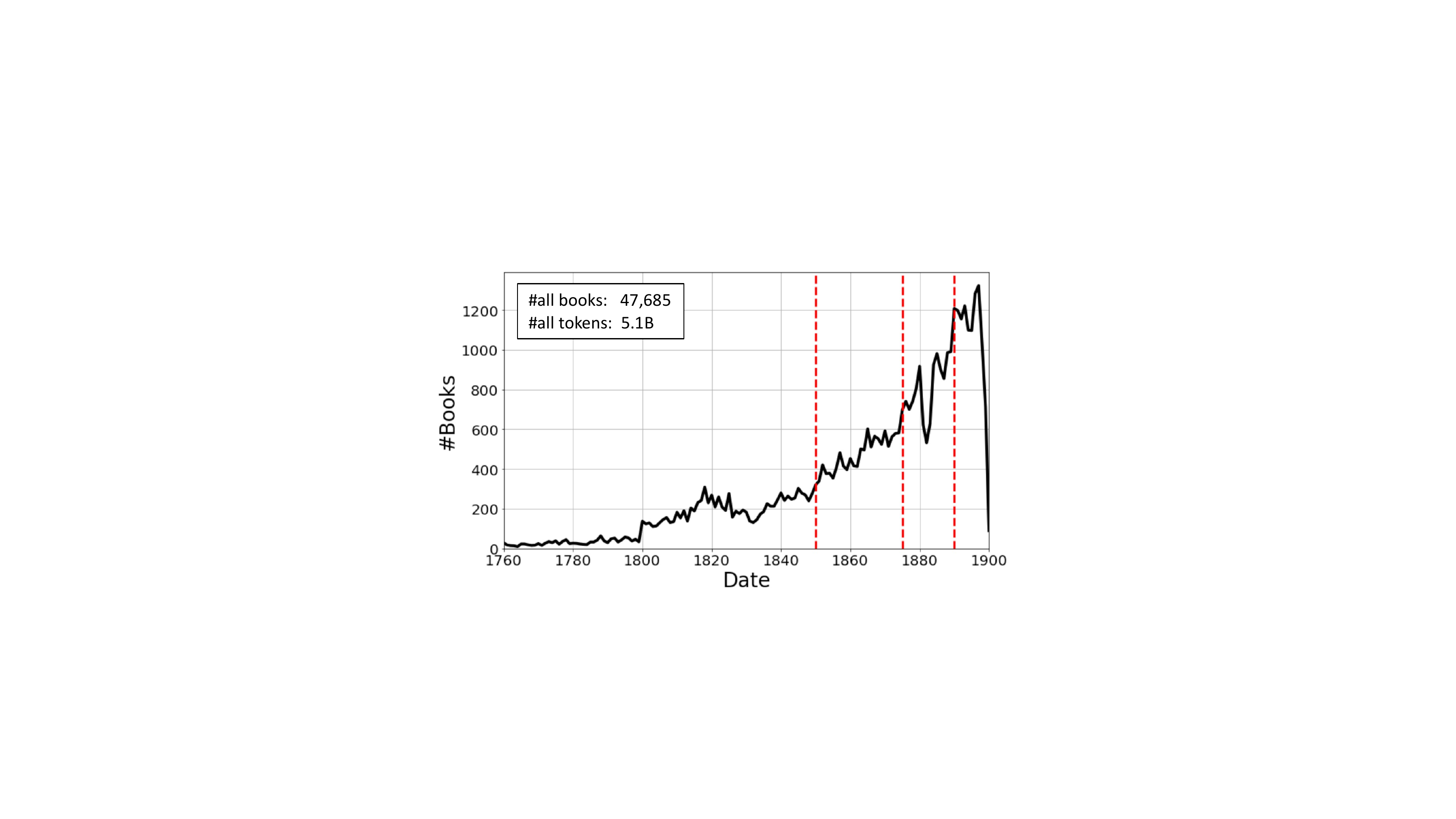}
	\end{center}
	\caption{Number of books by publication date. The preprocessed dataset has 47,685 books in English consisting of $\approx$5.1 billion tokens. The red vertical dashed lines mark the boundaries between the time periods we used to slice the dataset. See Section~\ref{sec:steps} for details.}
	\label{fig:num_books}
\end{figure}

The original collection consists of $\approx$48K digitized books in English, made openly available by the British Library in partnership with Microsoft, henceforth \textit{Microsoft British Library} Corpus  (MBL).
The digitized books are available as JSON files from the British Library web page.\footnote{\url{https://data.bl.uk/digbks/db14.html} (last access: 2021-05-24).}
Figure~\ref{fig:num_books} gives an overview of the number of books by year.
The bulk of the material is dated somewhere between 1800 and 1900, with the number of documents steeply rising at the end of the 19th century. 
Because all copyrights are cleared and the data are in the public domain, they have already become a popular resource for (digital) historians and literary scholars.\footnote{See for example the Contagion Project \url{https://cca.ucd.ie/contagion-project-british-library-corpus/} (last access: 2021-05-24).}
However, one notable issue with this collection (when used for historical research) is the somewhat opaque selection process of books: while the data provide decent coverage over the nineteenth century, the exact criteria for inclusion remain unclear and future work might profitably consider assessing the characteristics of this collection in more detail.

% ===================
\subsection{Steps} 
\label{sec:steps}

\paragraph{Preprocessing} Each book was minimally normalized: we converted the text to ASCII, fixed common punctuation errors, dehyphenated broken tokens, removed most punctuation and separated the remaining punctuation marks from tokens. While the large majority of books in the MBL corpus are written in English, the collection still contains a substantial amount of documents in other languages. Therefore, we filtered by English language, using \texttt{spaCy}'s language detector~\citep{spacysoftware}. Finally, we used \texttt{syntok}\footnote{\url{https://pypi.org/project/syntok} (last access: 2021-05-24).} to split the book into sentences and tokenize the text. This process resulted in one file per book where each line corresponded to a sentence with space-separated tokens.

\noindent \paragraph{Data selection} For each of the selected models, we trained an instance using the whole dataset (i.e.,~books from all over the 19th century; see Figure~\ref{fig:num_books}). For the word2vec and fastText models, we have also trained instances on text published before 1850. Moreover, for BERT, we have fine-tuned four model instances on different time slices, with data from before 1850, between 1850 and 1875, between 1875 and 1890, and between 1890 and 1900, each slice containing $\approx$1.3B tokens per period, except for 1890-1900, which included $\approx$1.1B tokens. While this periodization was largely motivated by the number of tokens, the different models (that resulted from the data partitioning) may enable historians to track cultural changes over the long 19th century.\footnote{For example, the pre-1850 dataset sets apart the first industrial revolution from later developments in Britain. Likewise, 1890-1900 is set off, especially in literary terms, by the emergence of `modernist' sensibilities and the questioning of class and gender hierarchies associated with the term `\textit{fin de si\`{e}cle}'.}

\paragraph{Word2vec and fastText} We trained the word2vec~\citep{mikolovEfficientEstimationWord2013} and fastText~\citep{bojanowski2016enriching} models as implemented in the Gensim library~\citep{rehurek2011gensim}. In addition to the preprocessing steps described above, we lowercased all tokens before training. For word2vec, we used the skip-gram architecture, which we trained for one epoch. We set the dimension of the word embedding vectors to 300 and removed tokens appearing less than 20 times. The same hyperparameters were used for training fastText models.

\paragraph{Flair} Flair is a character language model based on the Long short-term memory (LSTM) variant of recurrent neural networks~\citep{HochSchm97, akbik2019flair}. Even though less popular than the Transformers, it has been shown to obtain state-of-the-art results in Named Entity Recognition (NER).  We trained a character-level, forward-pass Flair language model on all the books in the MBL corpus for one epoch and sequence length of 250 characters (during training). We used the default character dictionary in Flair. The LSTM component had one layer and a hidden dimension of 2048.

\paragraph{BERT} To fine-tune BERT model instances, we started with a contemporary model: `BERT base uncased',\footnote{\url{https://github.com/google-research/bert} (last access: 2021-05-24).} hereinafter referred to as \textit{BERT-base} \citep{devlin-etal-2019-bert,Wolf2019HuggingFacesTS}. This instance was then fine-tuned on the earliest time period (i.e., books predating 1850). For the consecutive period (1850-1875), we used the pre-1850 language model instance as a starting point and continued fine-tuning with texts from the following period. This procedure of consecutive incremental fine-tuning was repeated for the other two time periods.  

We used the original BERT-base tokenizer as implemented by HuggingFace\footnote{\url{https://github.com/huggingface/transformers} (last access: 2021-05-24).}~\citep{Wolf2019HuggingFacesTS}. We did not train new tokenizers for each time period. This way, the resulting language model instances can be compared easily with no further processing or adjustments. The tokenized and lowercased sentences were fed to the language model fine-tuning tool in which only the masked language model (MLM) objective was optimized. We used a batch size of 5 per GPU and fine-tuned for 1 epoch over the books in each time-period. The choice of batch size was dictated by the available GPU memory (we used 4$\times$ NVIDIA Tesla K80 GPUs in parallel). Similar to the original BERT pre-training procedure, we used the Adam optimizer~\citep{kingma2014adam} with a learning rate of 0.0001, $\beta_{1} = 0.9$, $\beta_{2} = 0.999$ and $L_{2}$ weight decay of 0.01. In our fine-tuning procedure, we used a linear learning-rate warm-up over the first 2,000 steps. A dropout probability of 0.1 was applied in all layers.

\noindent \paragraph{Quality control} The quality of our language models was mainly evaluated on multiple downstream tasks. In \cite{van2020assessing}, we investigated the impact of OCR quality on the 19th-century word2vec model and showed how language models trained on large OCR’d corpora still yield robust word embedding vectors. The BERT models have been used in \cite{coll-ardanuy-etal-2020-living} and \cite{beelen-etal-2021-tsd}, where they generally improved the performance of various downstream tasks when the data of the experiments was contemporaneous to that of the language models, thereby confirming their quality via extrinsic evaluation.

% ===================
\section{Language model zoo}

\paragraph{Object name} \texttt{histLM}.

\paragraph{Format names and versions} The models are shared as ZIP files (one per model architecture). The directory structure is described in the \texttt{README.md} file.

\paragraph{Creation dates} 2020-01-31 to 2020-10-07.

\paragraph{Dataset creators} Kasra Hosseini, Kaspar Beelen and Mariona Coll Ardanuy (The Alan Turing Institute) preprocessed the text, created a database, trained and fine-tuned language models as described in this paper. Giovanni Colavizza (University of Amsterdam) initiated this work on historical language models. All authors contributed to planning and designing the experiments.

\paragraph{Language} The language models have been trained on 19th-century texts in English.

\paragraph{License} The models are released under open license CC BY 4.0, available at \url{https://creativecommons.org/licenses/by/4.0/legalcode}.

\paragraph{Repository name} All the language models are published in Zenodo at \url{http://doi.org/10.5281/zenodo.4782245}. We have also provided scripts to work with the language models, available on GitHub at \url{https://github.com/Living-with-machines/histLM}.

\paragraph{Publication date} 2021-05-23.

% ===================
\section{Reuse Potential}

Even though word2vec has been around almost a decade---an eternity in the fast-moving NLP ecosystem---the static word embeddings it produces persist as popular instruments, especially for interdisciplinary research~\citep{azarbonyad2017words,hengchen2019data}. The more recent fastText model extends on word2vec by using subword information. Contextualized language models have meant a breakthrough in NLP research (see e.g., \cite{smith2019contextual} for an overview), as they represent words in the contexts in which they appear, instead of conflating all senses, one of the main criticisms of static word embeddings. The potential of using such models for historical research is immense as they allow a more accurate context-dependent representation of meaning. 

Given that existing libraries such as Gensim, Flair, or Hugging Face provide convenient interfaces to work with these embeddings, we are confident that our historical models will serve the needs of a wide-variety of scholars, from NLP and data science to the humanities, for different tasks and research purposes, such as measuring how words change meaning over time, automatic OCR correction~\citep{hamalainen2019paft}, interactive query expansion\footnote{See for example the search tools provided by the Impresso interface \url{https://impresso-project.ch} (last access: 2021-05-24).} or, more generally, any research that involves diachronic language change~\citep{shoemark2019room}.

% ===================
\section*{Acknowledgements}

We are grateful to David Beavan (The Alan Turing Institute) and James Hetherington (University College London) for helping with the data access and research infrastructure. We thank the British Library for supplying digitised books.

% ===================
\section*{Funding Statement}
This work was supported by Living with Machines (AHRC grant AH/S01179X/1) and The Alan Turing Institute (EPSRC grant EP/N510129/1).

% ===================
\section*{Competing interests} 
The authors have no competing interests to declare.

% ===================
\theendnotes

% ===================
\bibliographystyle{johd}
\bibliography{bib}

\end{document}